%% file: main.tex
\documentclass[conference]{IEEEtran}

\usepackage{hyperref}       
\usepackage{url}            
\usepackage{booktabs}       
\usepackage{amsfonts}       
\usepackage{nicefrac}       
\usepackage{microtype}      
\usepackage{enumitem}
\usepackage{sidecap}

\usepackage{multicol}
\usepackage{amsmath}
\usepackage{graphicx}

\usepackage{algorithmic}
\usepackage{algorithm}

\usepackage{amssymb}
\usepackage{graphicx}
\usepackage{amssymb,amsmath,array}
\usepackage{amsmath,amsfonts,amsthm,bm} 

\usepackage{tikz}
\usetikzlibrary{arrows}
\usepackage{epstopdf}
\usepackage{times}

\usepackage{graphicx}
\usepackage{pgfgantt}

\usepackage{subfig}
\usepackage[graphicx]{realboxes}

\usepackage{amsmath}
\usepackage{graphicx}
\usepackage{bbm}
\usepackage{datetime}
\usepackage{color}
\definecolor{dblue}{rgb}{0,0,0.6}
\usepackage{amssymb}
\usepackage{relsize}
\usepackage{pifont}
\usepackage{multirow}

\usepackage{pifont}
\usepackage{multicol}
\usepackage{graphicx}
\usepackage{mathtools}
\usepackage{amssymb}
\usepackage{amsthm}
\usepackage{wrapfig,hyperref}
\hypersetup{colorlinks=true,linkcolor=blue,citecolor=red}

\newcommand\blfootnote[1]{%
  \begingroup
  \renewcommand\thefootnote{}\footnote{#1}%
  \addtocounter{footnote}{-1}%
  \endgroup
}

\usepackage[numbers]{natbib}
\usepackage{longtable}

\title{Flexible Online Representation Learning \\ Based on Similarity Matching}

\author{\IEEEauthorblockN{Shagesh Sridharan}
\IEEEauthorblockA{\textit{Physics and Astronomy} \\
\textit{Rutgers University}\\
New Brunswick, NJ, USA \\
shagesh.sridharan@rutgers.edu}
\and
\IEEEauthorblockN{Yanis Bahroun}
\IEEEauthorblockA{\textit{CCM, CCN } \\
\textit{Flatiron Institute, Simons Foundation}\\
New York, NY, USA \\
ybahroun@flatironinstitute.org}
\and
\IEEEauthorblockN{Anirvan M. Sengupta}
\IEEEauthorblockA{\textit{CCM, CCQ } \\
\textit{Flatiron Institute, Simons Foundation}\\
New York, NY, USA \\
anirvans.physics@gmail.com}
}

\setcitestyle{square}

\input{commands.tex}

\begin{document}

\maketitle

\begin{abstract}
Sparse high-dimensional representations are conducive to uncovering nontrivial structures in unsupervised exploration of data. Such a representation can deal with the dense connectivity in graphs relevant to community detection problems. However, sparse high-dimensional representations are capable of doing more, including manifold tiling and feature learning. Conventional algorithms optimize in the space of computationally intractable completely positive matrices or relax the problem to the space of doubly nonnegative matrices that scale with sample size in a way rendering them impractical for large data sets. Some of these methods also impose a row sum constraint,  such as double stochasticity. Row sum constraints have the added advantage of being shift-invariant, in the context of manifold tiling. Constraints on the row sum of output similarity matrices require nontrivial online learning rules. Addressing these needs, we propose a versatile online biologically plausible learning algorithm capable of learning sparse shift-invariant representations, useful for  clustering, manifold tiling, or sparse coding, depending on the data structure. 
%
%
\end{abstract}

\blfootnote{This manuscript is based on work originally accepted to IJCNN 2023 but not presented owing to visa issues.}

\begin{IEEEkeywords}
neural networks, manifold tiling, feature learning, community detection, SDP problems, online learning.
\end{IEEEkeywords}


\section{Introduction}

Recent developments in the machine learning and neural networks literature have shown a renewed interest in extracting structure by looking at alignment between input and embeddings representing the inputs \cite{nokland2016direct,pehlevan2019neuroscience}. Some of these models have shown to be useful, for example, for learning features \cite{pehlevan2014hebbian,bahroun2017online} or tiling manifolds \cite{tepper2018clustering,anirvan2018manifold}, or proposing an alternative to backpropagation \cite{nokland2019training,qin2021contrastive}. Interestingly, the objective functions for aligning the Gramian of inputs and the Gramian of embeddings  have long been of interest in the community detection literature \cite{amini2018semidefinite}. Community detection is an important problem in network analysis \cite{fortunato2010community,girvan2002community}. We connect it to the manifold tiling and feature learning literature.  We show that, depending upon the input dataset and choice of constraints, we can go from hard clustering to soft clustering for manifold-tiling \cite{anirvan2018manifold}, to a sparse coding regime for feature learning. The first two regimes lead to learning prototype learning, while the last one leads to a more distributed ``concept/atom''-based sparse coding representation in the style of dictionary learning \cite{olshausen2004sparse}.

The original offline, unrelaxed community detection problem corresponds to a clustering problem, closely related to the famous $k$-means algorithm \cite{macqueen1967some,lloyd1982least}. Indeed, it tries to find a solution in the set of completely positive (CP) matrices, which is computationally intractable although it is convex. As a result, various works have proposed relaxation that instead offers to solve a problem in the set of positive semidefinite (PSD) matrices \cite{amini2018semidefinite,awasthi2015relax,boumal2016non,peng2007approximating}. Such a solution seems a priori a good choice as it is now tractable but at a higher computational cost due to the semidefinite programming solvers. As a result, this apparent tractability is an illusion when dealing with large datasets. Instead, for large datasets, an online algorithm that can directly address the CP formulation is a more appealing alternative but, to the best of our knowledge, does not exist outside of this work. 

In particular, suppose we want to use a naive dot product kernel of the data without having to center the data, and we want a shift-invariant objective function. In that case, it is crucial to introduce a normalization over the rows of the output matrix of similarity, which has never been done online. 

To summarize, our contributions are the following. First, we propose to show that the objective function first introduced for community detection is a much more versatile problem than initially intended. Indeed, our objective function can learn sparse representations of the data or perform clustering, or tile manifolds, depending on the regime and the existence of the data structure. 
Second, we propose an online/streaming algorithm that works for large-scale datasets starting from the CP formulation of the problem without resorting to an SDP relaxation while enforcing a constraint that allows shift-invariance of the representation. 
Finally, we propose a neural implementation of our online algorithm with local learning rules, which is an essential criterion for building a biologically plausible learning algorithm. 
We test our algorithm on synthetic and image datasets and obtain competitive performance.  
\\ \\
\textbf{Notations.}
For positive integers $n$, $m$, let $\Real^n$ denote $n$-dimensional Euclidean space, and let $\Real^{n\times m}$ denote the set of $n \times m$ matrices equipped with the Frobenius norm $\| \cdot \|_F$. Boldface lowercase letters (e.g., $\vvec$) denote vectors, and boldface uppercase letters (e.g., $\M$) denote matrices. We let $\mathcal{S}^{p}_+$ denote the set of $p \times p$ positive semidefinite matrices. We let $\I_p$ denote the $p\times p$ identity matrix, and $\1_p$ the $p$-dimensional vector whose entries are equal to 1. 
For a set of inputs $\x_t \in \mathbb{R}^n$, $t = 1,\ldots,T$, we denote by $\X \in \mathbb{R}^{n \times T}$ the input matrix where the columns of the matrix are the input vectors. Similarly, the encoding/output matrix $\Y \in \mathbb{R}^{m \times T} = [\y_1,\ldots, \y_T ]$.

\section{Background and Related Work}\label{sec:background}

As mentioned earlier, two seemingly unrelated sets of problems have greatly inspired this work and are the subject of this section. On the one hand, the issue of community detection and, in particular, solving intractable representation alignment problems using SDP relaxations. 
The second set of problems is that of building biologically plausible learning algorithms that neural networks can implement and that can bring novel insights into neural computation.

\subsection{SDP Problems for Community Detection}

Community detection is one of the prominent problems in social network analysis. The exact recovery of community structure from networks generated by a Stochastic Block Model (SBM) \cite{holland1983stochastic} spurred the development of many algorithms \cite{zhao2017survey}. Newman and collaborators \cite{newman2004finding,newman2006modularity} study community detection based on a modularity measure that includes terms of the form $\tr( \mathbf{A}\Y^\top\Y)$ where $\mathbf{A}$ is the adjacency matrix of the graph and $\Y$ is a matrix whose columns are one-hot encodings of whether a sample belongs to a community or not. 

For an undirected graph with edge weight matrix $\D$, one could formulate the problem of maximizing $\tr (\D\Q)$, where $\Q$ is completely positive (CP), subject to some additional condition. Completely positive matrices are Gramians of some (entrywise) nonnegative matrix: $\Q=\Y^\top \Y$ with $\Y\ge0$. These formulations lead to CP relaxations of the original problem \cite{yazdanparast2017modularity}. One variant of the problem is
\begin{align}\label{eq:CP_clustering}
    \min_{\Q \in \textrm{CP}} - \Tr(\D\Q)  ~~ \text{s.t.}  ~~ \Q \1 = \1~~, \quad \Tr(\Q) = k. 
\end{align}
Unfortunately, these problems are NP-hard in the generic case \cite{brandes2006maximizing}. This led to further relaxation into doubly nonnegative matrices, namely, to nonnegative  positive semidefinite matrices $\Q$ \cite{cai2015robust,chen2014improved,chen2016statistical}.
The SDP problem closely related to \eqref{eq:CP_clustering} \cite{amini2018semidefinite}, is
\begin{align}\label{eq:SDP_clustering}
    \min_{\Q \in \mathcal{S}^T_{+}} - \Tr(\D\Q) ~~~ \text{s.t.} ~ \Q \ge 0 ~, ~ \Q 1 = 1 ~,~ \Tr(\Q) = k. 
\end{align}
Considerable study of this model shows it to be capable of discovering structure in a wide range of datasets \cite{kulis2007fast,awasthi2015relax}.
The condition $\Q\1 = \1$ is important for shift-invariance when we use a dot-product kernel for $\D$, as we will discuss later.

\subsection{Similarity-preserving neural networks}

In an effort to build biologically plausible neural networks from a normative approach, the authors of \cite{pehlevan2014hebbian,seung2017correlation} considered objective functions reminiscent of Eq.~\eqref{eq:SDP_clustering}, and of Classical Multi-Dimensional Scaling \cite{cox2000multidimensional}.
They departed from the standard reconstruction approach, which had led to the popular Oja's algorithms \cite{oja1989neural}. 
Indeed, existing models have led to non-local learning rules, contradicting the Hebbian principle of plasticity. 
The objective they considered is the following 
\begin{align}\label{eq:NSM_Orig}
   \argmin{\Y \in \Real_{+}^{m\times T} } -2 \Tr(\X^\top \X \Y^\top \Y) +  \Tr(\Y^\top \Y \Y^\top \Y).
\end{align}
The authors of \cite{pehlevan2014hebbian} noted that solving Eq.~\eqref{eq:NSM_Orig} directly necessitates storing $\Y\in \mathbb{R}_+^{m \times T}$ and is not a priori a natural choice for building online and biologically plausible algorithms. Indeed, the dimensionality of the similarity matrices increases with time $T$, which makes online learning difficult. However, they showed the superiority of this objective for building plausible neural networks (c.f. \cite{pehlevan2019neuroscience} for review), as shown in \cite{lipshutz2021biologically,bahroun2021normative}.

It is clear that the two objective functions Eq.~\eqref{eq:SDP_clustering}  and ~\eqref{eq:NSM_Orig} share a lot of similarities and have each their advantages. Indeed, in the SDP case, various constraints have been imposed that allow for more flexibility in the model but as a result, it is harder to solve online. On the other hand, similarity matching is easier to solve but remains more restrictive \cite{luther2022kernel}.


\section{Novel Objective Function and Min-max Formulation}
%

In this work, we propose to combine both of the frameworks mentioned in Sec.~\ref{sec:background}. Our goal is to propose a versatile tool for extracting information by either manifold-tiling, clustering, or feature learning. We further the objective function introduced by the authors of \cite{anirvan2018manifold} with additional constraints on the output Gram matrix. 
The objective function we consider is 
\begin{align}\label{eq:global-1}
&\Y^* = \argmin{\Y\geq 0} -\mathrm{Tr}(\X^\top \X \Y^\top \Y) 
\\& \mathrm{s.t.} \quad \Y^\top \Y \1_T = \gamma T \1_T, ~~  \mathrm{Tr}(\Y^\top \Y) \leq \beta T ~, \nonumber
\end{align}
where $\X^\top \X$ is the input Gram matrix corresponding to the input similarity matrix. The trace and row constraints in Eq.~\eqref{eq:global-1} are similar to that of the community detection problem Eq.~\eqref{eq:SDP_clustering}, but applied on the Gramian directly.

Notice that if each column of $\X$ is shifted by a vector $\mathbf{a}$ then $\X\to\X+\mathbf{a}\1_T^\top$, resulting in $\mathrm{Tr}(\X^\top \X \Y^\top \Y) \to \mathrm{Tr}(\X^\top \X \Y^\top \Y)+2\gamma T\mathrm{Tr}(\mathbf{a}^\top\X\1_T)+\gamma T^2||\mathbf{a}||^2$. 
The optimization is thus not affected by the shift.

Importantly, we will not resort to an SDP relaxation but will solve the problem in the set of CP matrices directly.

%
%
\subsection{A Min-max Formulation} 
In this section, we introduce a formulation of our objective function, Eq.~\eqref{eq:global-1}, that is amenable to online learning (Sec.~\ref{sec:online_implementation}) and leads to a neural implementation (Sec.~\ref{subsec:neural}). For the sake of clarity, we separate the procedure into four steps. 
\\
\\
\textbf{Step 1: Incorporate the constraints.} We add the constraints, into the objective Eq.~\eqref{eq:global-1} using two sets of Lagrange multipliers, $\alpha_t$ and $\lambda$ as follows
\begin{align}\label{eq:NSM_2}
    \min_{\Y \geq 0} \max_{\alpha_t,\lambda \geq 0}  - & \frac{1}{2T^2}\Tr\left( \X^\top \X \Y^\top \Y \right) + \tfrac{\lambda}{2T} \left( \sum_t \|\y_t\|^2 - \beta T \right) \nonumber \\ 
    & + \tfrac{1}{T^2}\sum_t\alpha_t \left( \y_t^\top \sum_{\tilde{t}} \y_{\tilde{t}}  - \gamma T \right)  .
\end{align}
\textbf{Step 2: Introduce auxiliary variables.} We now introduce a set of auxiliary variables, denoted by $\W$, which will simplify the first term in Eq.~\eqref{eq:NSM_2} as 
\begin{align}\label{eq:W_opt}
    - \frac{1}{2T^2} \Tr (\X^\top\X \Y^\top \Y) = \min_{\W\in\mathbb{R}^{m\times n}} & - \frac{1}{T}\Tr(\X^\top \W^\top \Y ) \nonumber \\
    &+ \frac{1}{2} \Tr(\W^\top\W) ~.
\end{align}
\textbf{Step 3: Auxiliary variables to relax the constraints.} We simplify the last term in our objective function, Eq.~\eqref{eq:NSM_2} by introducing the couplet $(\uvec_+,\uvec_{-})$ as
\begin{align*}
    & \frac{1}{T^2}\left(\sum_t \alpha_t \y_t^\top \right) \left( \sum_{\tilde{t}} \y_{\tilde{t}} \right) = \max_{\uvec_{+}} \min_{\uvec_{-}} -\frac{\|\uvec_{+}\|^2}{4} +  \frac{\|\uvec_{-}\|^2}{4} \\ 
    & \qquad \qquad + \frac{\uvec_{+}^\top}{2T} \left(\sum_t (1 + \alpha_t) \y_t\right)  
     + \frac{\uvec_{-}^\top}{2T} \left(\sum_t (1 - \alpha_t) \y_t\right).
\end{align*}
\textbf{Step 4: Combined objective function.} Now that we have introduced the necessary auxiliary variables, we obtain the following combined objective function
\begin{align}\label{eq:minmax_offline}
    & \min_{\y_t \geq 0} \max_{\alpha_t,\lambda \geq 0,\uvec_{+}} \min_{\uvec_{-},\W} - \frac{1}{T}\sum_{t} \x_t^\top \W^\top \y_t + \frac{1}{2} \Tr\left(\W^\top \W\right) \nonumber \\
    & - \sum_t\frac{\gamma}{T} \alpha_t + \frac{\lambda}{2T} \sum_t \left( \|\y_t\|^2 - \beta T \right) -\frac{\|{\uvec_{+}}\|^2}{4} + \frac{\|{\uvec_{-}}\|^2}{4}  \nonumber \\ 
     & \quad + \frac{\uvec_{+}^\top}{2T} \left(\sum_t (1 + \alpha_t) \y_t\right) + \frac{{\uvec_{-}^\top}}{2T} \left(\sum_t (1 - \alpha_t) \y_t\right).
\end{align}

The novel combined objective can seem a priori more involved than the original one, however, it has the important property of having decoupled samples at different times, i.e., $t \neq \tilde{t}$, which initially made the problem hard to solve online.

%
%
%
%
%
%
\begin{algorithm}[t]
  \caption{Online algorithm for problem \eqref{eq:global-1}.}
  \label{alg:online}
\begin{algorithmic}
  \STATE {\bfseries input} data $\{\x_1,\dots,\x_T\}$; dimension $n$  
  \STATE {\bfseries output} $\{\y_1,\dots,\y_T\}$; dimension $m$ 
  \STATE {\bfseries initialize} the matrix $\W$, and vectors $\uvec_{+}$, and $\uvec_{-}$.
  \FOR{$t=1, 2,\dots,T $}
  \STATE {Taking $\x_t$ as input run the neural dynamics defined by equations \eqref{eq:y_dynamics} and \eqref{eq:alpha_dynamics} until convergence.} 
 \STATE {Update $\W$, $\uvec_{+}$ and $\uvec_{-}$ as in equations \eqref{eq:w_update}, and \eqref{eq:u+_update} using the previously computed $\y_t$ and $\alpha_t$.}
  \ENDFOR
\end{algorithmic}
\end{algorithm}

\section{Online Algorithm from Min-max Problem}\label{sec:online_implementation}

In this section, we present an online algorithm for solving CP problem \eqref{eq:global-1} from min-max objective \eqref{eq:minmax_offline} and its biologically plausible neural network implementation using neural dynamics and synaptic plasticity rules.

\subsection{Derivation of the online algorithm}

We now solve the min-max objective \eqref{eq:minmax_offline} in the online setting. At each time step $t$, we first solve the problem for $\y_t$ and $\alpha_t$ by projected gradient descent-ascent as 
\small
\begin{align}
  \y_t &   \gets  [\y_t + \delta ( \W \x_t - \lambda \y_t - \tfrac{\uvec_{+}}{2} (1 + \alpha_t) - \tfrac{\uvec_{-}}{2} (1 - \alpha_t))]_+ \label{eq:y_dynamics}
  \\
    \alpha_t & \gets \alpha_t + \delta \left( \frac{1}{2}\left( \uvec_{+}^\top - \uvec_{-}^\top\right) \y_t - \gamma   \right) ~, \label{eq:alpha_dynamics}
\end{align}
\normalsize
where $[\vvec_t]_+ = \max(\vvec_t,0)$. 
In the second step, we update $\W$, $\uvec_{+}$, $\uvec_{-}$ by gradient descent-ascent as 
\begin{align}
\W & \gets \W  + \eta \left( \y_t\x_t^\top - \W \right) ~, \label{eq:w_update} \\
\uvec_{\pm} & \gets \uvec_{\pm} + \eta\left[ \left( \alpha_t \pm 1 \right){\y_t} - \uvec_{\pm} \right] ~. \label{eq:u+_update} 
\end{align}
This yields our online algorithm (Algorithm~\ref{alg:online}) and the neural network implementing it, which we describe in more detail in the following section (Sec.~\ref{subsec:neural}).

\subsection{Neural Interpretation}\label{subsec:neural}

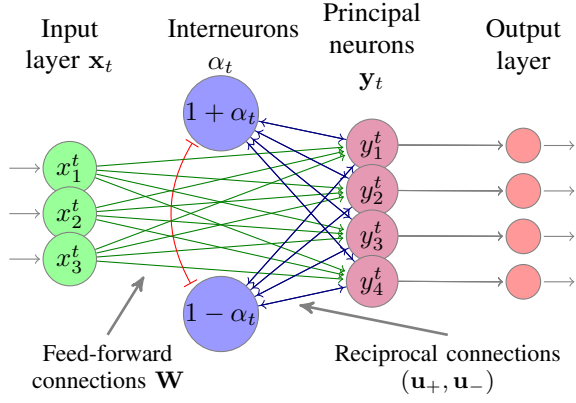
\begin{figure}[!t]
\centering
\def\hlayersep{0cm}
\def\layersep{2cm}
\tikzstyle{information text}=[text badly centered,font=\small,text width=3cm]
\begin{tikzpicture}[shorten >=1pt,->,draw=black!50, node distance=\layersep]
    \tikzstyle{every pin edge}=[<-,shorten <=1pt]
    \tikzstyle{neuron}=[circle,draw,fill=black!100,minimum size=13pt,inner sep=0pt]
    \tikzstyle{input neuron}=[neuron, fill=green!40];
    \tikzstyle{output neuron}=[neuron, fill=red!40];
    \tikzstyle{latent neuron}=[neuron, fill=blue!40];
    \tikzstyle{hidden neuron}=[neuron, fill=purple!40];
    \tikzstyle{annot} = [text width=4em, text centered]

 \begin{scope}[>=stealth', line width=1pt]
        \draw[->] (-1.5,-2.6) node[below, information text]
            {Feed-forward connections $\W$ } -- (-1,-1.8);
        \draw[->] (3,-2.6) node[below,information text]
            {Reciprocal  connections $(\uvec_+,\uvec_{-})$} -- (1,-2.2);
    \end{scope}
    \foreach \name / \y in {1,...,3}
        \node[input neuron,pin={[pin edge={<-}]left:}, left of=(I-\name)] (I-\name) at (0,0.2-\y*0.6) {$~x^t_\y~$};

    \path[yshift=2.825cm]
    node[latent neuron] (L-1) at (\hlayersep,-1*2.5 cm) {$1 + \alpha_t$};
    \path[yshift=2.675cm]
    node[latent neuron] (L-2) at (\hlayersep,-2*2.5 cm) {$1 - \alpha_t$};
    
    \foreach \name / \y in {1,...,4}
        \path[yshift=0.5cm]
            node[hidden neuron] (H-\name) at (\layersep,0.2-\y*0.6 cm) {$~y^t_\y~$};

    \foreach \name / \y in {1,...,4}
        \path[yshift=0.5cm]
            node[output neuron,pin={[pin edge={->}]right:}, right of=(O-\name)] (O-\name) at (\layersep,-\y*0.6 cm) {};    	

    \foreach \source in {1,...,3}
        \foreach \dest in {1,...,4}
            \path (I-\source) edge[green!50!black] (H-\dest);

    \foreach \source in {1,...,2}
	    \foreach \dest in {1,...,4}
     	   \path (H-\dest) edge[blue!50!black] (L-\source);

    \foreach \source in {1,...,2}
	    \foreach \dest in {1,...,4}
     	   \path (L-\source) edge[blue!50!black] (H-\dest);

    \foreach \source in {1,...,4}
	    \foreach \dest in {1,...,4}
     	   \path (H-\source) edge (O-\source);

    \path ($(L-2) - (0.4cm,-0.4cm)$) edge[|-|, color=red, style={bend left}] ($(L-1) - (0.36cm,0.36cm)$);
     	   
    \node[annot,above of=H-1, node distance=1.3cm] (hl) {Principal neurons $\y_t$};
    \node[annot,left of=hl] (ll) {Interneurons $\alpha_t$};
    \node[annot,left of=ll] {Input layer $\x_t$};
    \node[annot,right of=hl] {Output layer};
\end{tikzpicture}
\caption{\label{fig:Network} A single-layer neural network with interneurons, implementing the dynamics and update rules from Eq.~\eqref{eq:y_dynamics}. 
}
\end{figure}

Using the dynamics Eqs.~\eqref{eq:y_dynamics}-\eqref{eq:alpha_dynamics} and the update rules in Eqs.~\eqref{eq:w_update}-\eqref{eq:u+_update} our online algorithm has a biologically plausible network interpretation. The network architecture is represented in Figure \ref{fig:Network}.

Each component of the output vector $\y_t$ is represented in the activity of principal neurons, which receives input projected onto a synaptic weight vector encoded in the rows of $\W$. Instead, $1-\alpha_t$ and $1+\alpha_t$, represent the activity of two interneurons connected with the principal neurons, with reciprocal connections encoded in $\uvec_{+}$ and $\uvec_{-}$. The two interneurons are constrained such that their total sum of activity is constant, equal to 2.  
These fast dynamics stabilize the principal and inter-neurons before passing the output downstream. 
The update rule for $\W$, Eq.~\eqref{eq:w_update}, is a local Hebbian synaptic plasticity rule. It is also the case for the update rules of $\uvec_+$ and $\uvec_{-}$. These update rules happen on a slower time scale than the neural dynamics.

In summary, our online model operates in two steps. First, it computes a representation for each new input $\x_t$ by encoding it as post-synaptic activity $\y_t$ coupled with the latent variable $\alpha_t$. Second, it updates the synaptic weights between nodes using local Hebbian rules that require only the current pre- and post-synaptic neuronal activities.
%

%
%
%
%
%
%

\section{Experiments and Results}\label{sec:exp}

In the following experiments, we evaluate the performance of our algorithms on two different tasks: \textbf{(1)} manifold tiling (Sec.~\ref{sec:Task_Tiling}) and \textbf{(2)} feature learning (Sec.~\ref{sec:Task_Feature}). 

\subsection{Task 1: Online Manifold Tiling}\label{sec:Task_Tiling}

In this section, we demonstrate the capacity of our online algorithm to learn to tile manifolds with constraints while generating output neurons with localized receptive fields that respond to a small neighborhood of stimulus space \cite{okeefe1978hippocampus}.
%
%

\begin{figure*}[!ht]
    \includegraphics[width=.95\textwidth]{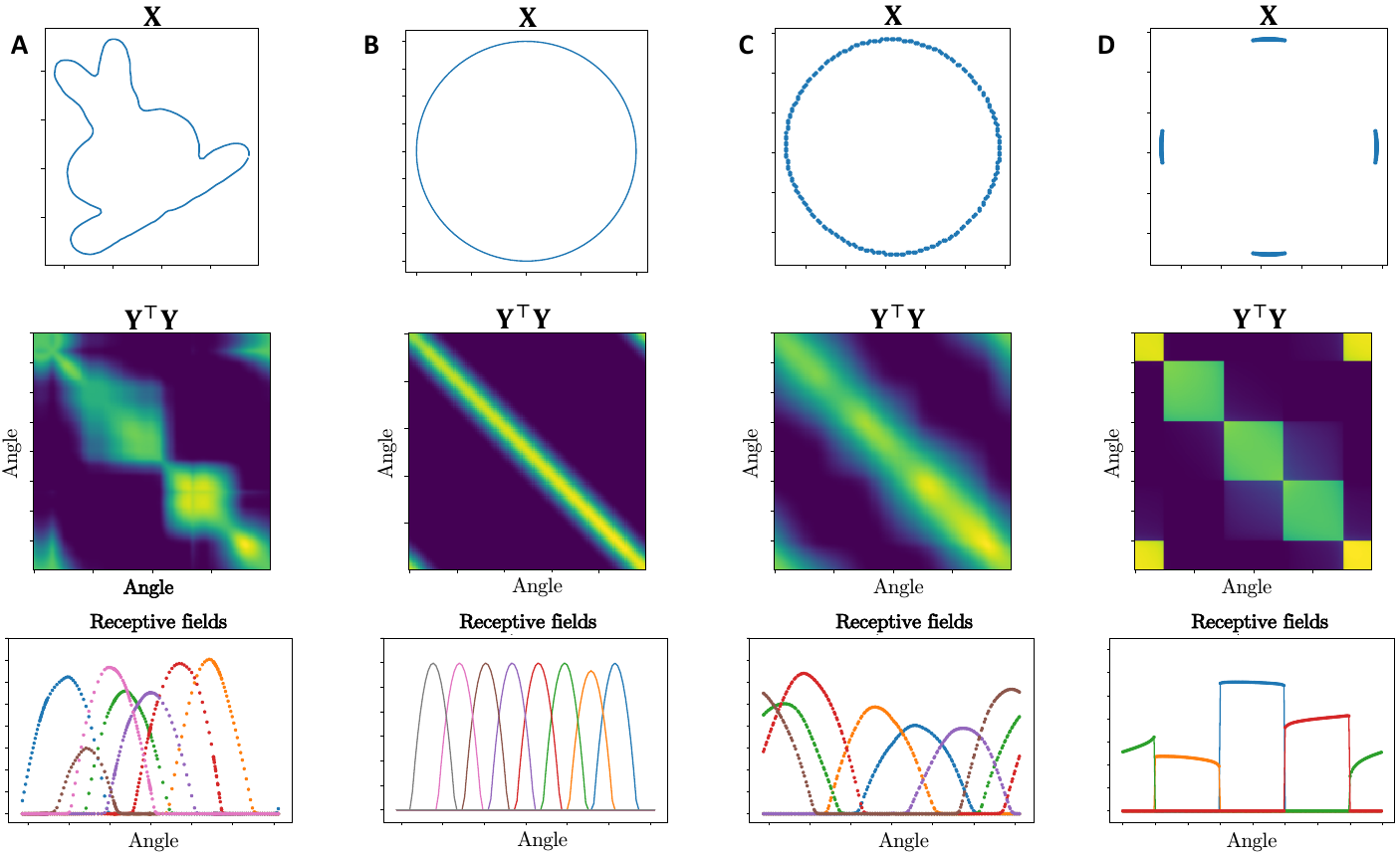}
    \caption{Our algorithm is able to tile various manifolds despite the non-uniform curvature and non-uniform density of points. The top middle and bottom rows, respectively represent the inputs $\X$, the output Gram matrix learned by the model, $\Y^\top \Y$, and the receptive fields associated with each output neuron. 
    } \label{fig:tiling_synth}
\end{figure*}

%
%
\textbf{Datasets and parameters.} 
%
We assess performance on four synthetic datasets (Fig.~\ref{fig:tiling_synth} top panels). The input data vectors ${\x_1,\ldots, \x_T}$ are sampled points from either Bunny-like shapes or 2D circular rings, as illustrated in Fig.~\ref{fig:tiling_synth}\textbf{A}-\textbf{B}. Fig.~\ref{fig:tiling_synth}\textbf{C} displays non-uniformly sampled inputs on the ring, while Fig.~\ref{fig:tiling_synth}\textbf{D} shows four clusters equally spaced on the rim. We set the number of output neurons to $m=100$ and the trace constraint parameter $\beta = 1$ for all datasets. In Fig.~\ref{fig:tiling_synth}\textbf{A}-\textbf{C}-\textbf{D}, we use a row constraint parameter $\gamma = 0.25$, while in Fig.~\ref{fig:tiling_synth}\textbf{B}, we set $\gamma = 0.05$.
%

%
%

%
%
\textbf{Results.} 
In the middle panels of Fig.~\ref{fig:tiling_synth}, we show the learned matrix of similarity, $\Y^\top \Y$, as a function of the angle. In Fig.~\ref{fig:tiling_synth}\textbf{A}-\textbf{D}, we have abused the notation of angle as parameterizing the data points in the clockwise direction around its center. The $\Y^\top \Y$ panels show that the learned Gramians are nonnegative and high rank, unlike the rank-2 input Gramians. The bottom panels show the receptive fields learned by our model for the different neurons. It shows that the receptive fields are localized and tile the manifolds as expected theoretically. In short, our model performs online manifold tiling while enforcing the row and trace constraints confirming our theoretical derivation of the algorithm. 
%
%

\subsection{Task 2: Online Feature Learning}\label{sec:Task_Feature}
We assess our algorithm's ability to learn useful features for image classification by employing a pipeline from \cite{coates2011analysis} that evaluates unsupervised learning methods.
%
In short, the features learned by the model are used as input to an SVM to test their effectiveness in classification tasks. 
%
%
%

\begin{figure*}[!t]
\centering
    \includegraphics[width=.95\textwidth]{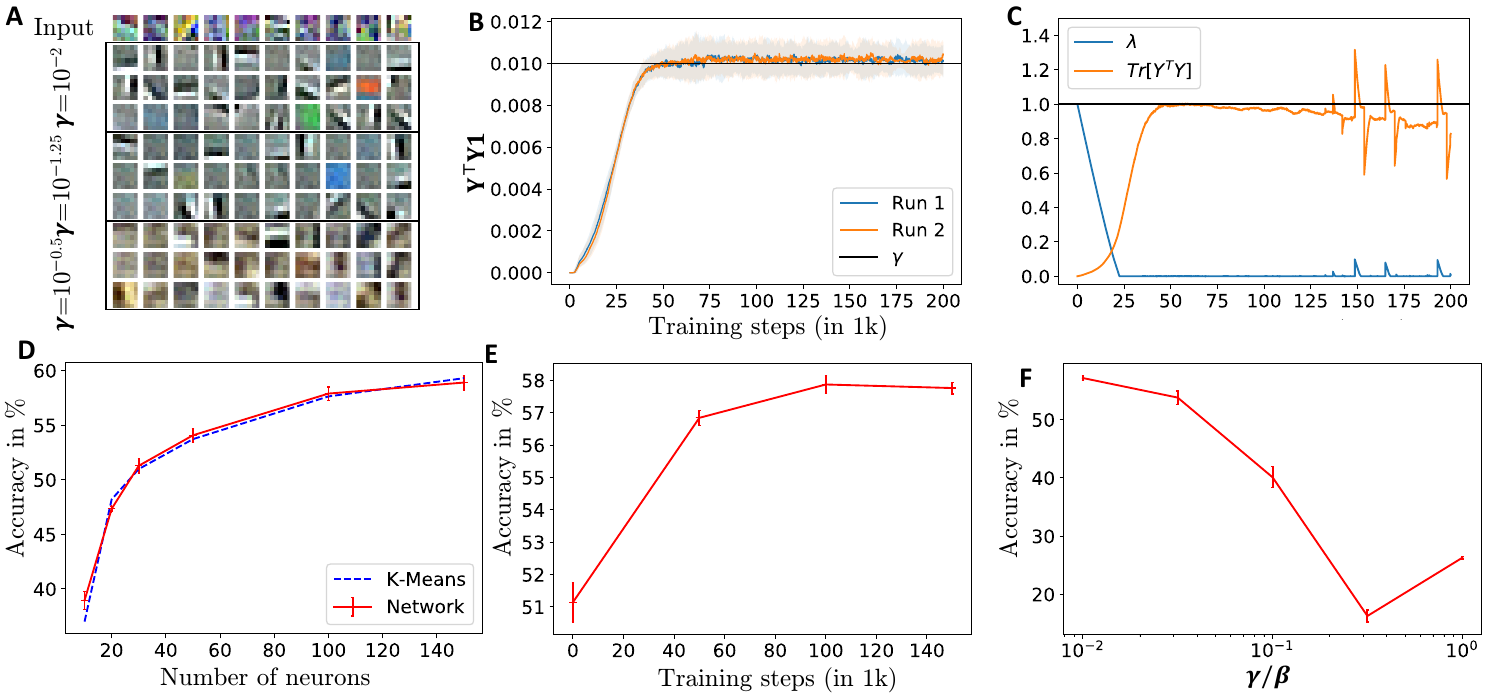}
    \caption{Our algorithm is able to learn features that are useful for classification tasks. \textbf{A} shows a $n=100$ features learned by our model online. \textbf{B} shows the row sum of the Gramian, which is constrained to $\gamma T = 10^{-2}$ The results are averaged over five trials and show the mean and std deviations.  Instead, \textbf{C} shows $\Tr(\Y^\top\Y)$, and $\alpha_t$. Finally, \textbf{D-E-F} respectively shows the accuracy of our model as a function of the number of output neurons, $m$, the number of training steps, $T$, and the ratio $\gamma/\beta$. 
    } \label{fig:feature_learning}
\end{figure*}

\textbf{Datasets and parameters.} 
We evaluate our model using the CIFAR-10 dataset \cite{krizhevsky2009learning}, consisting of 50k training images and 10k test images. 
Pipeline parameters include input patch size $n$ and feature number $m$.
Following \cite{coates2011analysis}, the model learns from patches of images rather than the entire image. 
Additionally, we investigate the influence of the sparsity ratio $\gamma/\beta$ (introduced in Eq.~\eqref{eq:global-1}) on model performance.
%
%
%
%

\textbf{Results.} 
We show in Fig.~\ref{fig:feature_learning}\textbf{A} the features learned by our neural network when trained on whitened input patches, of size $n= 6\times 6 \times 3$, and three different $\gamma$ values. 
For $\gamma = 10^{-2}$, the sparsest of the three configurations, the features are sharp, localized, and resemble edge filters, reminiscent of preferred features of the primary visual cortex \cite{olshausen1996emergence}. 
With increasing $\gamma$, i.e., decreasing sparsity, the features deteriorate in sharpness and exhibit degeneracy with increasing color specificity. 

We show in Fig.~\ref{fig:feature_learning}\textbf{B-C} respectively that the row constraint, $\Y^\top \Y \1_T = \gamma T \1_T$, and the trace constraint, $\Tr(\Y^\top \Y) \leq \beta T$, are fulfilled by our model online. 
In Fig.~\ref{fig:feature_learning}\textbf{D}, we show that the performance of our model increases as a function of the number of output neurons, $m$, and that it shows competitive performance with the popular $k$-means algorithm as in \cite{coates2011analysis}. 
Fig.~\ref{fig:feature_learning}\textbf{E} suggests that, indeed, the features learned by the network over time help the system improve in accuracy.

Finally, we show  in  Fig.~\ref{fig:feature_learning}\textbf{F} the influence of sparsity on classification accuracy. It shows that a decrease in sparsity induces a decrease in accuracy. This result is consistent with the tiling intuition that increasing the ``diameter'' yields a lower-rank embedding.

%
%
%
%
%
\section{Conclusion}

In this work, we introduced a novel constrained CP problem that can be solved online and can serve as a versatile tool for extracting information from data. We also propose a biologically plausible neural network implementation of the online algorithm. 
%
%
Our algorithms represent manifolds with high-dimensional vectors of assignment indices to centroids, demonstrating localized receptive fields instead of low-dimensional embedding, unlike most existing algorithms. 
Additionally, our model learns discriminative representations of image datasets online for classification tasks, which proves beneficial in further computation. 
These findings demonstrate the versatility of similarity-preserving neural networks for information extraction. 

\section*{Acknowledgment}

This work was supported in part by Simons Foundation grant SF 626323 to Anirvan M. Sengupta.

\bibliographystyle{IEEEtran}
\bibliography{bib_file}



\end{document}

%% file: commands.tex
\DeclareMathOperator{\tr}{Tr}
\DeclareMathOperator{\Tr}{Tr}

\newcommand{\argmin}[1]{\underset{#1}{\operatorname{arg}\,\operatorname{min}}\;}

\newcommand{\uvec}{{\bf u}}
\newcommand{\vvec}{{\bf v}}

\newcommand{\x}{{\bf x}}
\newcommand{\y}{{\bf y}}

\newcommand{\D}{{\bf D}}

\newcommand{\I}{{\bf I}}

\newcommand{\M}{{\bf M}}

\newcommand{\Q}{{\bf Q}}

\newcommand{\W}{{\bf W}}
\newcommand{\X}{{\bf X}}
\newcommand{\Y}{{\bf Y}}

\newcommand{\1}{{\bf 1}}

\newcommand{\Real}{{\mathbb R}}

%% file: bib_file.bib
@article{oja1989neural,
  title={Neural Networks, Principal Components, and Sub-spaces},
  author={Erkki Oja},
  journal={International Journal of Neural Systems},
  volume={Vol. 01, No. 01},
  pages={61--68},
  year={1989}
}

@article{olshausen1996emergence,
  title={Emergence of simple-cell receptive field properties by learning a sparse code for natural images},
  author={Bruno A. Olshausen and David J. Field },
  journal={Nature},
  volume={381},
  pages={607--609},
  year={1996}
}

@article{anirvan2018manifold,
  title={Manifold-tiling localized receptive fields are optimal in similarity-preserving neural networks},
  author={Sengupta, Anirvan and Pehlevan, Cengiz and Tepper, Mariano and Genkin, Alexander and Chklovskii, Dmitri},
  journal={Advances in Neural Information Processing Systems},
  volume={31},
  year={2018}
}

@book{cox2000multidimensional,
    title = {Multidimensional Scaling},
    author = {Trevor F. Cox and Mike Cox},
    isbn = {1584880945},
    year = {2000},
    publisher = {Chapman and Hall/CRC}
}

@inproceedings{pehlevan2014hebbian,
  title={A {H}ebbian/anti-{H}ebbian network for online sparse dictionary learning derived from symmetric matrix factorization},
  author={Hu, Tao and Pehlevan, Cengiz and Chklovskii, Dmitri B},
  booktitle={2014 48th Asilomar Conference on Signals, Systems and Computers},
  pages={613--619},
  year={2014},
  organization={IEEE}
}

@article{amini2018semidefinite,
  title={On semidefinite relaxations for the block model},
  author={Amini, Arash A and Levina, Elizaveta},
  journal={The Annals of Statistics},
  volume={46},
  number={1},
  pages={149--179},
  year={2018},
  publisher={Institute of Mathematical Statistics}
}

@inproceedings{awasthi2015relax,
  title={Relax, no need to round: Integrality of clustering formulations},
  author={Awasthi, Pranjal and Bandeira, Afonso S and Charikar, Moses and Krishnaswamy, Ravishankar and Villar, Soledad and Ward, Rachel},
  booktitle={Proceedings of the 2015 Conference on Innovations in Theoretical Computer Science},
  pages={191--200},
  year={2015}
}

@article{peng2007approximating,
  title={Approximating k-means-type clustering via semidefinite programming},
  author={Peng, Jiming and Wei, Yu},
  journal={SIAM Journal on Optimization},
  volume={18},
  number={1},
  pages={186--205},
  year={2007},
  publisher={SIAM}
}

@article{boumal2016non,
  title={The non-convex {B}urer-{M}onteiro approach works on smooth semidefinite programs},
  author={Boumal, Nicolas and Voroninski, Vlad and Bandeira, Afonso},
  journal={Advances in Neural Information Processing Systems},
  volume={29},
  year={2016}
}

@article{fortunato2010community,
  title={Community detection in graphs},
  author={Fortunato, Santo},
  journal={Physics reports},
  volume={486},
  number={3-5},
  pages={75--174},
  year={2010},
  publisher={Elsevier}
}

@article{girvan2002community,
  title={Community structure in social and biological networks},
  author={Girvan, Michelle and Newman, Mark EJ},
  journal={Proceedings of the national academy of sciences},
  volume={99},
  number={12},
  pages={7821--7826},
  year={2002},
  publisher={National Academy of Sciences}
}

@inproceedings{bahroun2017online,
  title={Online representation learning with single and multi-layer {H}ebbian networks for image classification},
  author={Bahroun, Yanis and Soltoggio, Andrea},
  booktitle={International Conference on Artificial Neural Networks},
  pages={354--363},
  year={2017},
  organization={Springer}
}

@inproceedings{coates2011analysis,
  title={An analysis of single-layer networks in unsupervised feature learning},
  author={Coates, Adam and Ng, Andrew and Lee, Honglak},
  booktitle={Proceedings of the Fourteenth International Conference on Artificial Intelligence and Statistics},
  pages={215--223},
  year={2011},
  organization={JMLR Workshop and Conference Proceedings}
}

@article{tepper2018clustering,
  title={Clustering is semidefinitely not that hard: {N}onnegative {SDP} for manifold disentangling},
  author={Tepper, Mariano and Sengupta, Anirvan M and Chklovskii, Dmitri},
  journal={The Journal of Machine Learning Research},
  volume={19},
  number={1},
  pages={3208--3237},
  year={2018},
  publisher={JMLR. org}
}

@inproceedings{kulis2007fast,
  title={Fast low-rank semidefinite programming for embedding and clustering},
  author={Kulis, Brian and Surendran, Arun C and Platt, John C},
  booktitle={Artificial Intelligence and Statistics},
  pages={235--242},
  year={2007},
  organization={PMLR}
}

@article{lloyd1982least,
  title={Least squares quantization in {PCM}},
  author={Lloyd, Stuart},
  journal={IEEE Transactions on Information Theory},
  volume={28},
  number={2},
  pages={129--137},
  year={1982},
  publisher={IEEE}
}

@inproceedings{macqueen1967some,
  title={Some methods for classification and analysis of multivariate observations},
  author={MacQueen, James},
  booktitle={Proceedings of the fifth Berkeley symposium on mathematical statistics and probability},
  volume={1},
  pages={281--297},
  year={1967},
  organization={Oakland, CA, USA}
}

@article{nokland2016direct,
  title={Direct feedback alignment provides learning in deep neural networks},
  author={N{\o}kland, Arild},
  journal={Advances in Neural Information Processing Systems},
  volume={29},
  year={2016}
}

@misc{krizhevsky2009learning,
  title={Learning multiple layers of features from tiny images},
  author={Krizhevsky, Alex and Hinton, Geoffrey},
  year={2009},
  publisher={Citeseer}
}

@article{pehlevan2019neuroscience,
  title={Neuroscience-inspired online unsupervised learning algorithms: Artificial neural networks},
  author={Pehlevan, Cengiz and Chklovskii, Dmitri B},
  journal={IEEE Signal Processing Magazine},
  volume={36},
  number={6},
  pages={88--96},
  year={2019},
  publisher={IEEE}
}

@article{bahroun2021normative,
  title={A Normative and Biologically Plausible Algorithm for {I}ndependent {C}omponent {A}nalysis},
  author={Bahroun, Yanis and Chklovskii, Dmitri and Sengupta, Anirvan},
  journal={Advances in Neural Information Processing Systems},
  volume={34},
  year={2021}
}

@article{lipshutz2021biologically,
  title={A biologically plausible neural network for multichannel canonical correlation analysis},
  author={Lipshutz, David and Bahroun, Yanis and Golkar, Siavash and Sengupta, Anirvan M and Chklovskii, Dmitri B},
  journal={Neural Computation},
  volume={33},
  number={9},
  pages={2309--2352},
  year={2021},
  publisher={MIT Press One Rogers Street, Cambridge, MA 02142-1209, USA journals-info~…}
}

@book{okeefe1978hippocampus,
  title={The hippocampus as a cognitive map},
  author={O'Keefe, John and Nadel, Lynn},
  year={1978},
  publisher={Oxford University Press}
}

@article{qin2021contrastive,
  title={Contrastive similarity matching for supervised learning},
  author={Qin, Shanshan and Mudur, Nayantara and Pehlevan, Cengiz},
  journal={Neural Computation},
  volume={33},
  number={5},
  pages={1300--1328},
  year={2021},
  publisher={MIT Press One Rogers Street, Cambridge, MA 02142-1209, USA journals-info~…}
}

@inproceedings{nokland2019training,
  title={Training neural networks with local error signals},
  author={N{\o}kland, Arild and Eidnes, Lars Hiller},
  booktitle={International Conference on Machine Learning},
  pages={4839--4850},
  year={2019},
  organization={PMLR}
}

@article{holland1983stochastic,
  title={Stochastic blockmodels: First steps},
  author={Holland, Paul W and Laskey, Kathryn Blackmond and Leinhardt, Samuel},
  journal={Social Networks},
  volume={5},
  number={2},
  pages={109--137},
  year={1983},
  publisher={Elsevier}
}

@article{zhao2017survey,
  title={A survey on theoretical advances of community detection in networks},
  author={Zhao, Yunpeng},
  journal={Wiley Interdisciplinary Reviews: Computational Statistics},
  volume={9},
  number={5},
  pages={e1403},
  year={2017},
  publisher={Wiley Online Library}
}

@article{newman2004finding,
  title={Finding and evaluating community structure in networks},
  author={Newman, Mark EJ and Girvan, Michelle},
  journal={Physical Review E},
  volume={69},
  number={2},
  pages={026113},
  year={2004},
  publisher={APS}
}

@article{newman2006modularity,
  title={Modularity and community structure in networks},
  author={Newman, Mark EJ},
  journal={Proceedings of the national academy of sciences},
  volume={103},
  number={23},
  pages={8577--8582},
  year={2006},
  publisher={National Academy of Sciences}
}

@article{yazdanparast2017modularity,
  title={Modularity maximization using completely positive programming},
  author={Yazdanparast, Sakineh and Havens, Timothy C},
  journal={Physica A: Statistical Mechanics and its Applications},
  volume={471},
  pages={20--32},
  year={2017},
  publisher={Elsevier}
}

@article{brandes2006maximizing,
  title={Maximizing modularity is hard},
  author={Brandes, Ulrik and Delling, Daniel and Gaertler, Marco and G{\"o}rke, Robert and Hoefer, Martin and Nikoloski, Zoran and Wagner, Dorothea},
  journal={arXiv preprint physics/0608255},
  year={2006}
}

@article{cai2015robust,
  title={Robust and computationally feasible community detection in the presence of arbitrary outlier nodes},
  author={Cai, T Tony and Li, Xiaodong},
  journal={The Annals of Statistics},
  volume={43},
  number={3},
  pages={1027--1059},
  year={2015},
  publisher={Institute of Mathematical Statistics}
}

@article{chen2014improved,
  title={Improved graph clustering},
  author={Chen, Yudong and Sanghavi, Sujay and Xu, Huan},
  journal={IEEE Transactions on Information Theory},
  volume={60},
  number={10},
  pages={6440--6455},
  year={2014},
  publisher={IEEE}
}

@article{chen2016statistical,
  title={Statistical-computational tradeoffs in planted problems and submatrix localization with a growing number of clusters and submatrices},
  author={Chen, Yudong and Xu, Jiaming},
  journal={The Journal of Machine Learning Research},
  volume={17},
  number={1},
  pages={882--938},
  year={2016},
  publisher={JMLR. org}
}

@article{olshausen2004sparse,
  title={Sparse coding of sensory inputs},
  author={Olshausen, Bruno A and Field, David J},
  journal={Current opinion in neurobiology},
  volume={14},
  number={4},
  pages={481--487},
  year={2004},
  publisher={Elsevier}
}

@article{seung2017correlation,
  title={A correlation game for unsupervised learning yields computational interpretations of Hebbian excitation, anti-Hebbian inhibition, and synapse elimination},
  author={Seung, H Sebastian and Zung, Jonathan},
  journal={arXiv preprint arXiv:1704.00646},
  year={2017}
}

@article{luther2022kernel,
  title={Kernel similarity matching with Hebbian Networks},
  author={Luther, Kyle and Seung, Sebastian},
  journal={Advances in Neural Information Processing Systems},
  volume={35},
  pages={2282--2293},
  year={2022}
}
